\documentclass[11pt,a4paper]{article}
\usepackage[hyperref]{emnlp2020}
\usepackage{times}
\usepackage{latexsym}
\usepackage{graphicx}
\graphicspath{ {./images/} }
\usepackage{xcolor}

\usepackage{microtype}

\aclfinalcopy 

\title{Enhancing the Identification of Cyberbullying through Participant Roles}

 \author{Gathika Ratnayaka$^1$,  Thushari Atapattu$^2$, Mahen Herath$^1$, Georgia Zhang$^2$ \and Katrina Falkner$^2$ \\ $^1$Department of Computer Science \& Engineering, University of Moratuwa, Katubedda, Sri Lanka \\ $^2$School of Computer Science, The University of Adelaide, Adelaide, Australia \\ email: thushari.atapattu@adelaide.edu.au}

\date{}

\begin{document}
\maketitle

\begin{abstract}
Cyberbullying is a prevalent social problem that inflicts detrimental consequences to the health and safety of victims such as psychological distress, anti-social behaviour, and suicide. 
The automation of cyberbullying detection is a recent but widely researched problem, with current research having a strong focus on a binary classification of bullying versus non-bullying. This paper proposes a novel approach to enhancing cyberbullying detection through role modeling. We utilise a dataset from ASKfm to perform multi-class classification to detect participant roles (e.g. victim, harasser). Our preliminary results demonstrate promising performance including 0.83 and 0.76 of F1-score for cyberbullying and role classification respectively, outperforming baselines.
\end{abstract}

\section{Introduction}

The surge of Internet and social media has led to the unprecedented social crisis of cyberbullying, particularly among adolescents. It can lead to various damaging consequences on the health and safety of victims, such as feelings of isolation, depression, and suicide. Cyberbullying is the \textit{repetitive use of aggressive language among peers, with the intention to harm others through digital media} \cite{Rosa2019}.  
 Despite the illegality of harassing others, most social media platforms are susceptible to cyberbullying due to the openness and anonymisation of platforms. Research conducted by \citet{Patchin2019} indicates that cyberbullying victimisation rates have approximately doubled between the years 2007 and 2019. Adolescents, minorities (e.g. refugees, LGBTQI) and women are among common targets of cyberbullying. The sheer amount of cyberbullying-related incidents vastly exceeds the capacity of manual detection and demands the need to develop technology to effectively and automatically detect this.

\begin{figure}
    \includegraphics[width=8cm, height=3cm]{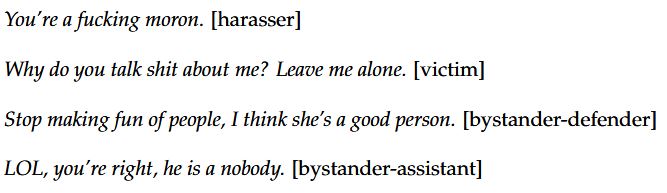}
    \caption{An excerpt from a cyberbullying episode \cite{VanHeeReport2015}}
    \label{fig:bullying}
\end{figure}

The development of automated models to detect cyberbullying is a widely researched problem in recent years, with current research focusing on classifying posts as bullying or non-bullying \cite{Rosa2019,algardi2016,Salawu2020}. One of the fundamental gaps in current research is that \textit{all texts from all users are treated equally} without differentiating who has authored bullying and who has been targeted.These models provide a temporary solution by filtering offensive contents. Bullies often find novel ways to bypass technology such as incorporating implicit and subtle forms of language (e.g. sarcasm) and pseudo profiles.  Identifying the roles of authors and targets introduces a novel approach to enable more information-rich models and to foster precise detection. A small number of recent studies focus on cyberbullying-related \textit{'participant roles'} (e.g. bully, victim, bystander) (see Figure \ref{fig:bullying}) \cite{Vanhee2018,Xu2012,Jacob2020}. 

Motivated by this idea, our work focuses on two tasks, 1) detecting cyberbullying as a binary classification problem, and 2) detecting participant roles as a multi-class classification problem. We build upon  previous role identification research and the AMiCA dataset proposed by \citet{Vanhee2018}. 
\section{Related Works}
In addition to modeling bullying and non-bullying content as a binary classification task \cite{Rosa2019,algardi2016,Salawu2020}, several research studies focus on participant role identification \cite{Salawu2020,Vanhee2018,Xu2012} within the cyberbullying context. \citet{Xu2012} defined 8 roles - \textit{bully, victim, bystander, assistant, defender, reporter, accuser} and \textit{reinforcer}, based on the theoretical framework of \citet{Salmivalli2010}. The majority of previous studies addressing role identification incorporate user- (e.g., age, gender, location) and social network-based features (e.g., number of followers, network centrality). Although these features have demonstrated a tendency to increase classification performance \cite{Huang2014,Vivek2016}, relying on user and network features is logistically challenging in real-world application due to the creation of pseudo profiles and ethical restrictions imposed by platforms. 
Alternatively, lexical and semantic features (e.g., subjectivity lexicons, character n-grams, topic models, profanity word lists, and named entities) of participants' posts are considered in few research studies \cite{Vanhee2018,Xu2012}. 

Our research aims to automatically identify cyberbullying and roles are based on supervised learning mechanisms that utilizes pretrained language models  and advanced contextual embedding techniques. Therefore, such mechanisms will mitigate the need for rule-based approaches and will also minimize the requirement for creating task-specific feature extraction mechanisms. 

\section{Model Description}
This study focuses on two tasks 1) detecting cyberbullying as a binary classification problem, and 2) detecting cyberbullying-related participant roles as a multi-class classification problem.

\subsection{Cyberbullying classification}
\label{ssec:modelcyber}
Instead of building new models, we extend an ensemble model originally designed by the authors \cite{Herath2020} for SemEval-2020 Task on offensive language identification \cite{Zampieri2020}, to classify posts in the current dataset. The reused ensemble model \cite{Herath2020} was built using three single classifiers, each based on DistilBERT \cite{Sanh2019}, a lighter, faster version of BERT \cite{devlin2018bert}. Each of the single classifiers A, B, and C was trained on a Twitter dataset containing Tweets annotated as offensive ('OFF') or non-offensive('NOT') posts. Models A and B were trained on imbalanced sets of Twitter data where the majority class instance was OFF and NOT respectively. Model C was trained using a balanced subset of Tweets which were assigned opposing class labels by the models A and B. 

Each classifier was trained using a learning rate of 5e-5 and a batch size of 32 for 2 epochs. A voting scheme was then used to combine the single models and build an ensemble model. If the biased classifiers A and B agreed upon a label for a given data instance, we assigned it that particular label. If the predictions from the biased classifiers were different, we assigned the data instance the prediction from the model C. This ensemble model achieved 0.906 of F1 score on the evaluation dataset of OffensEval challenge \cite{Zampieri2020}. 

\subsection{Role classification}
\label{ssec:modelrole}
\begin{figure}
    \includegraphics[width=8cm, height=4cm]{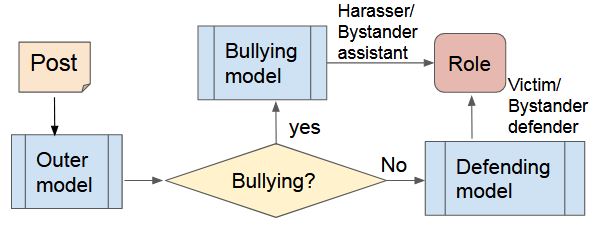}
    \caption{Overview of the Ensemble model}
    \label{fig:ensemble}
\end{figure}

According to a theoretical framework developed by \citet{Salmivalli2010} and 
the annotation guide by \citet{VanHeeReport2015}, ‘bystander assistant’ also engages in bullying while helping or encouraging the ‘harasser'. Similarly, ‘bystander defender’ helps the ‘victims’ to defend themselves from the harassment. Therefore, we consider 'bystander assistant' as a role which contributes to bullying. Accordingly, we categorise the posts of harassers and bystander assistants in AMiCA dataset into a category called ‘bullying’ and victim and bystander defender’s posts into a category called ‘defending’. Then, we divide the posts in each category into the roles as shown in Figure \ref{fig:ensemble}. The final ensemble model contains 3 sub models as follows,

\begin{enumerate}
    \item \textbf{Outer Model}: Classifies a post as Bullying or Defending
    \item \textbf{Bullying Model}: Classifies a post as 'Harasser' or 'Bystander assistant'
    \item \textbf{Defending Model}: Classifies a post as 'Victim' or 'Bystander defender'
\end{enumerate}

\begin{figure}
    \includegraphics[width=6cm, height=8cm]{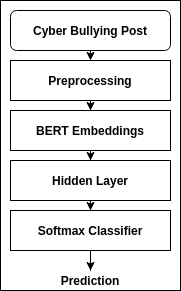}
    \caption{Model architecture}
    \label{fig:bert}
\end{figure}

Each of these models have the same model architecture, that consists of a pre-trained BERT embedding layer, hidden neural layer and a softmax output layer (Figure \ref{fig:bert}). In order to extract  BERT embeddings, ‘bert-based uncased’ model \cite{devlin2018bert} used. As discussed in section \ref{sec:results}, each model was experimented with different sampling strategies and cost functions to obtain optimal performance.

\section{Methods}
\label{sec:methods}
Our research is guided by two tasks, which focus on evaluating the performance of models that could classify whether a given post is, 
\begin{enumerate}
    \item cyberbullying-related or not, and
    \item if cyberbullying-related, predicting the role of the user who authored that post.
\end{enumerate}

\subsection{Dataset}
AMiCA dataset contains data collected from the social networking site ASKfm\footnote{https://ask.fm/} by \citet{Vanhee2018} in April and October, 2013. ASKfm is very popular among adolescents and has increasingly been used for cyberbullying \cite{Kao2019}. 
We used the English dataset, where posts are annotated and presented in chronological order within their original conversation (see Figure \ref{fig:bullying}). AMiCA dataset is annotated by linguists using BRAT\footnote{https://brat.nlplab.org/}, a web-based tool for text annotation, and considers the following four roles.
\begin{itemize}
    \item \textbf{Harasser}: person who initiates the harassment
    \item \textbf{Victim}: person who is harassed
    \item \textbf{Bystander defender}: person who helps the victim and discourages the harasser from continuing his actions
    \item \textbf{Bystander assistant}: person who does not initiate, but takes part in the actions of the harasser.
\end{itemize}

\begin{figure}
     \includegraphics[width=8cm, height=3cm]{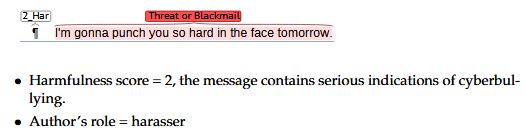}
     \caption{An example of BRAT annotation \cite{VanHeeReport2015}}
     \label{fig:brat}
\end{figure}

 Figure \ref{fig:brat} shows the annotation mechanism where ‘2\_Har’ refers that the author’s role is ‘harasser’ while the harmfulness score is 2. 

At post-level, the harmfulness of a post is scaled from 0 (no harm) to 2 (severely harmful).  We merge harmfulness scores 1 and 2 together (e.g. 1\_victim, 2\_victim as 'victim') to increase training examples for each cyberbullying role. The cyberbullying class contained 5,380 instances (Harasser - 3,576, Victim - 1,356, Bystander assistant - 24, Bystander defender - 424). AMiCA dataset also provides annotations of cyberbullying-related textual categories such as threat, insult, curse. This study does not focus on those annotations during our model development.


\citet{Vanhee2018} have used 10\% of the data as the hold-out test set. However, their hold-out is not publicly available. Therefore, in this study, we perform 10-fold cross validation while having 10\% of the dataset as the test set in each fold. In order to maintain a similar data distribution ratio among the classes and to make sure that test set of one fold is mutually exclusive with the test sets of other folds, we use the ‘StratifiedKFold’ method in the Scikit-Learner. 


\subsection{Data preprocessing and balancing}
In order to minimise the noise of ASKfm posts, we performed some pre-processing steps such as replacing slang words and abbreviations \footnote{https://floatcode.wordpress.com/2015/11/28/internet-slang-dataset/} and decoding emoticons\footnote{https://github.com/carpedm20/emoji} in addition to standard data pre-processing steps (e.g. removal of punctuations) while fine-tuning BERT \cite{devlin2018bert}.

Before feeding the posts into the models, we performed more preprocessing steps such as converting to lower case, tokenisation using the bert-tokenizer, and special token additions (adding [CLS] and [SEP] tokens to appropriate positions to perform BERT based sequence classification). 


\section{Results and Discussion}
\label{sec:results}
\textbf{Evaluation metric.} To evaluate our models and compare the performance with baselines, we use metrics similar to \citet{Vanhee2018}: 1) \textbf{F1-score:} The harmonic mean of precision and recall and 2) \textbf{Error rate:} 1- recall of the class.\\


\noindent\textbf{Baseline.} We use the best system of \citet{Vanhee2018} as our baseline to compare our models. This baseline used feature combinations such as subjectivity lexicons, character n-grams, term lists, and topic models.



\subsection{Evaluation of cyberbullying classification}
As discussed in section \ref{ssec:modelcyber}, our cyberbullying classification experiments extended an ensemble model (refer as ‘OffensEval ensemble’ hereafter) based on DistilBERT developed by authors for SemEval 2020 challenge \cite{Herath2020}. To test the performance of OffensEval ensemble on ASKfm dataset, we constructed three test datasets. Each test dataset consisted of 10,872 non-bullying posts randomly sampled from the non-cyberbullying class and all the 5,380 posts belonging to the cyberbullying class. The class distribution in test datasets was selected such that it would be compatible with \citet{Vanhee2018}. The averaged performance using three test sets is presented in Table \ref{tab:cyberbullying} along with the baselines.

According to the results, our OffensEval ensemble model outperforms the best system of \citet{Vanhee2018} by a margin of 0.2 (F1 score). Since present results were obtained by evaluating a pre-built model for a separate task, in our future works, we expect to improve our performance through fine-tuning our previous model on AMiCA dataset. Further, the presence of obscene slang words in non-cyberbullying posts could have led to some of the false positives. A sample of examples in this category is provided in section \ref{ssec:roleresults}. The presence of very short posts with 'chat-related slang words (e.g., \textit{Fgt, No to the woah hoe})' the model has not seen during the training could have led to some of the false negatives.

\begin{table}
\begin{tabular}{lccc}
\hline
\textbf{Model} & \textbf{F1} & \textbf{P} & \textbf{R}\\ 
\hline
OffensEval ensemble & \textbf{0.83} & \textbf{0.84} & \textbf{0.82} \\
\citet{Vanhee2018} & 0.64 & 0.74 & 0.56 \\\hline


\end{tabular}
\caption{Hold-out scores of cyberbullying classification}
\label{tab:cyberbullying}
\end{table}

\subsection{Evaluation of role classification}
\label{ssec:roleresults}
Table \ref{tab:rolemodels} demonstrates the 10-fold cross-validation results of our role classification models. As discussed in section \ref{ssec:modelrole}, we created the BERT-based ‘outer model’ to classify posts into two classes - bullying and defending. At the initial experiments, we obtained low recall for ‘defending’ class mainly due to the class imbalance in the dataset. To overcome this drawback, we have carried out experiments with different techniques such as weighted random sampling and weighted cross-entropy loss (as cost function). Based on the results of our experiments, weighted random sampling was used when training the outer model as it has shown considerable improvement in performance. Weighted random sampling is an sampling technique that attempts to maintain an approximately equal distribution of data instances among classes in a batch while training. 

Our BERT-based ‘defending model’ demonstrated promising performance including 0.93 of weighted F1 score and 0.96 (victim class) and 0.86 (bystander defender class) of F1 score (Table \ref{tab:rolemodels}). Our BERT-based ‘bullying model’ was not successful in classifying bystander assistants. We have experimented several strategies to improve the performance of bystander assistant detection such as choosing different training samples, limiting the number of instances taken from 'Harasser' class (100, 500) when training the 'Bullying' model, using weighted random sampling to under sample the harasser class while oversampling the bystander assistant class in order to keep the distribution among two classes at a ratio near to 1:1. However, these strategies failed to enable the 'Bullying' model or the overall ensemble model to detect bystander assistant class properly. Based on these experiments, we assume that the issue of the bystander assistant being classified as a harasser may not be due to class imbalance, however, based on the fact that examples in both classes have the overlapping language (see sample posts of 'bystander assistant' below).

\textit{"[..] wanna kill him? let's do it together"}

\textit{"[..] she's a massive sl*t! I agree with you [..] I'm on your side"}

While training each of the three models (Outer, Bullying, Defending), batch size of 8 was used with a maximum sequence length of 256 characters. Cross entropy loss was used as the cost function and stochastic gradient descent with a learning rate of \(2 \times 10^{-5}\) was used as the optimizer.

As shown in the Table \ref{tab:rolemodels}, our BERT-based ‘ensemble model’ has achieved ‘good’ performance (weighted F1-score is 0.76) except in the classes - victim and bystander assistant. According to the confusion matrix of ensemble model, most misclassified instances are related to victims being classified as harassers. An error analysis of misclassified posts revealed that bullying language widely overlaps with victims when victims use swear words to respond the harasser. These posts increase the difficulty for models to detect victims and require efforts in future research to develop effective models that can handle aggressive victims. A sample of posts where victims have aggressively responded to harassers is shown below.

\textit{"[..] whoever is saying that sh*t that its me needs to cut your sh*t out you need to shut the f*** up [..]"}

\textit{"and you're living proof that abortion should be legal"}

\begin{table}
\begin{tabular}{llllll}
\hline
\textbf{Model} & \textbf{WF} & \textbf{Class} & \textbf{P} & \textbf{R} & \textbf{F1}   \\ 
\hline
Outer & 0.78 & Bully & 0.84 & 0.83 & 0.83   \\
 & & Defend & 0.66 & 0.67 & 0.67   \\
Bullying & 0.99 & Harasser  & 0.99 & 1.00 & 1.00   \\
 & & B.assist & 0.20 & 0.05 & 0.08  \\
Defending & 0.93 & Victim & 0.95 & 0.96 & 0.95   \\ 
 & & B.defend & 0.86 & 0.84 & 0.85  \\ 
 Ensemble & 0.76 & Harasser & 0.84 & 0.82 & 0.83  \\
  &  & Victim & 0.59 & 0.61 & 0.60  \\
  & & B.assist & 0.00 & 0.00 & 0.00  \\
 & & B.defend & 0.68 & 0.73 & 0.70  \\ \hline
\end{tabular}
\caption{10-fold cross-validation scores of our models; WF: Weight F1}
\label{tab:rolemodels}
\end{table}

The comparison of our role classification model with the baselines is restricted since \citet{Vanhee2018} do not report cross-validation results\footnote{https://doi.org/10.1371/journal.pone.0203794.t009}, However, if the 'error rates' are compared using our 10-fold cross-validation results with their hold-out results, our model outperforms the baseline by 0.26 and 0.11 of 'error rate' in harasser and victim classes respectively. Both the models were not able to detect bystander assistant successfully (i.e. error rate is 1). The baseline outperforms us by 0.01 (error rate) in the bystander defender class. \citet{Vanhee2018} reported that error rates often being lowest for the profanity baseline, confirming that it performs well in terms of recall, however, precision is also an important metric to be considered. In our future work, we intend to further improving recall of each role class while stabilizing good precision. 



\section{Conclusions}
This paper proposes an approach to classify cyberbullying and associated roles (e.g., harasser, victim) as a novel contribution to enhance automated cyberbullying detection. Cyberbullying is a growing social problem that inflicts  detrimental impacts on online users. The identification of roles is a valuable contribution to future research as it can prompt closer monitoring of bullies and implicitly help victims through potential prevention. Currently, our approaches to identifying cyberbullying related roles focus only on individual posts on a forum. In our future work, we aim to expand this further by considering an entire discussion and the discourse relationships between the posts within the considered discussion. This will enable us to get a better understanding of the roles played by different users in a discussion. Moreover, we intend to integrate cyberbullying and role classification as a single model and optimise performance further to provide an effective solution to the cyberbullying problem.

\section*{Acknowledgments}

Authors would like to acknowledge the researchers on the AMiCA project for sharing the dataset.

\bibliographystyle{acl_natbib}
\bibliography{emnlp2020}

\begin{thebibliography}{16}
\expandafter\ifx\csname natexlab\endcsname\relax\def\natexlab#1{#1}\fi

\bibitem[{Al-garadi et~al.(2016)Al-garadi, Varathan, and Ravana}]{algardi2016}
Mohammed~Ali Al-garadi, Kasturi~Dewi Varathan, and Sri~Devi Ravana. 2016.
\newblock \href {https://doi.org/https://doi.org/10.1016/j.chb.2016.05.051}
  {Cybercrime detection in online communications: The experimental case of
  cyberbullying detection in the twitter network}.
\newblock \emph{Computers in Human Behavior}, 63:433 -- 443.

\bibitem[{Devlin et~al.(2018)Devlin, Chang, Lee, and
  Toutanova}]{devlin2018bert}
Jacob Devlin, Ming-Wei Chang, Kenton Lee, and Kristina Toutanova. 2018.
\newblock \href {http://arxiv.org/abs/1810.04805} {Bert: Pre-training of deep
  bidirectional transformers for language understanding}.

\bibitem[{Herath et~al.(2020)Herath, Atapattu, Dung, Treude, and
  Falkner}]{Herath2020}
Mahen Herath, Thushari Atapattu, Hoang Dung, Christoph Treude, and Katrina
  Falkner. 2020.
\newblock {AdelaideCyC at SemEval-2020 Task 12: Ensemble of Classifiers for
  Offensive Language Detection in Social Media}.
\newblock In \emph{Proceedings of SemEval}.

\bibitem[{Huang et~al.(2014)Huang, Singh, and Atrey}]{Huang2014}
Qianjia Huang, Vivek~Kumar Singh, and Pradeep~Kumar Atrey. 2014.
\newblock \href {https://doi.org/10.1145/2661126.2661133} {Cyber bullying
  detection using social and textual analysis}.
\newblock In \emph{Proceedings of the 3rd International Workshop on
  Socially-Aware Multimedia}, page 3–6. Association for Computing Machinery.

\bibitem[{Jacobs et~al.(2020)Jacobs, Van~Hee, and Hoste}]{Jacob2020}
Gilles Jacobs, Cynthia Van~Hee, and Véronique Hoste. 2020.
\newblock Automatic classification of participant roles in cyberbullying: Can
  we detect victims, bullies, and bystanders in social media text?
\newblock \emph{Natural Language Engineering}.
\newblock In-print.

\bibitem[{Kao et~al.(2019)Kao, Yan, Huang, Bartley, Hosseinmardi, and
  Ferrara}]{Kao2019}
Hsien-Te Kao, Shen Yan, Di~Huang, Nathan Bartley, Homa Hosseinmardi, and Emilio
  Ferrara. 2019.
\newblock \href {https://doi.org/10.1145/3308560.3316505} {Understanding
  cyberbullying on instagram and ask.fm via social role detection}.
\newblock In \emph{Companion Proceedings of The 2019 World Wide Web
  Conference}, page 183–188. Association for Computing Machinery.

\bibitem[{Patchin and Hinduja(2019)}]{Patchin2019}
J.~Patchin and S.~Hinduja. 2019.
\newblock \href {www.cyberbullying.org} {Lifetime cyberbullying victimization
  rates}.

\bibitem[{Rosa et~al.(2019)Rosa, Pereira, Ribeiro, Ferreira, Carvalho,
  Oliveira, Coheur, Paulino, {Veiga Simão}, and Trancoso}]{Rosa2019}
H.~Rosa, N.~Pereira, R.~Ribeiro, P.~C. Ferreira, J.~P. Carvalho, S.~Oliveira,
  L.~Coheur, P.~Paulino, A.~M. {Veiga Simão}, and I.~Trancoso. 2019.
\newblock \href {https://doi.org/https://doi.org/10.1016/j.chb.2018.12.021}
  {Automatic cyberbullying detection: A systematic review}.
\newblock \emph{Computers in Human Behavior}, 93:333 -- 345.

\bibitem[{Salawu et~al.(2020)Salawu, He, and Lumsden}]{Salawu2020}
S.~Salawu, Y.~He, and J.~Lumsden. 2020.
\newblock \href {https://doi.org/10.1109/TAFFC.2017.2761757} {Approaches to
  automated detection of cyberbullying: A survey}.
\newblock \emph{IEEE Transactions on Affective Computing}, 11(01):3--24.

\bibitem[{Salmivalli(2010)}]{Salmivalli2010}
Christina Salmivalli. 2010.
\newblock \href {https://doi.org/https://doi.org/10.1016/j.avb.2009.08.007}
  {Bullying and the peer group: A review}.
\newblock \emph{Aggression and Violent Behavior}, 15(2):112 -- 120.
\newblock Special Issue on Group Processes and Aggression.

\bibitem[{Sanh et~al.(2019)Sanh, Debut, Chaumond, and Wolf}]{Sanh2019}
Victor Sanh, Lysandre Debut, Julien Chaumond, and Thomas Wolf. 2019.
\newblock \href {http://arxiv.org/abs/1910.01108} {Distilbert, a distilled
  version of bert: smaller, faster, cheaper and lighter}.

\bibitem[{Singh et~al.(2016)Singh, Huang, and Atrey}]{Vivek2016}
Vivek~K. Singh, Qianjia Huang, and Pradeep~K. Atrey. 2016.
\newblock Cyberbullying detection using probabilistic socio-textual information
  fusion.
\newblock In \emph{Proceedings of the 2016 IEEE/ACM International Conference on
  Advances in Social Networks Analysis and Mining}, page 884–887.

\bibitem[{Van~Hee et~al.(2018)Van~Hee, Jacobs, Emmery, Desmet, Lefever,
  Verhoeven, De~Pauw, Daelemans, and Hoste}]{Vanhee2018}
Cynthia Van~Hee, Gilles Jacobs, Chris Emmery, Bart Desmet, Els Lefever, Ben
  Verhoeven, Guy De~Pauw, Walter Daelemans, and Véronique Hoste. 2018.
\newblock \href {https://doi.org/10.1371/journal.pone.0203794} {Automatic
  detection of cyberbullying in social media text}.
\newblock \emph{PLOS ONE}, 13(10):1--22.

\bibitem[{Van~Hee et~al.(2015)Van~Hee, Verhoeven, Lefever, De~Pauw, Daelemans,
  and Hoste}]{VanHeeReport2015}
Cynthia Van~Hee, Ben Verhoeven, Els Lefever, Guy De~Pauw, Walter Daelemans, and
  Véronique Hoste. 2015.
\newblock {Guidelines for the Fine-Grained Analysis of Cyberbullying, version
  1.0. LT3,}.
\newblock Technical report, Language and Translation Technology Team Ghent
  University.

\bibitem[{Xu et~al.(2012)Xu, Jun, Zhu, and Bellmore}]{Xu2012}
Jun-Ming Xu, Kwang-Sung Jun, Xiaojin Zhu, and Amy Bellmore. 2012.
\newblock Learning from bullying traces in social media.
\newblock In \emph{Proceedings of the 2012 Conference of the North American
  Chapter of the Association for Computational Linguistics: Human Language
  Technologies}, page 656–666. Association for Computational Linguistics.

\bibitem[{Zampieri et~al.(2020)Zampieri, Nakov, Rosenthal, Atanasova,
  Karadzhov, Mubarak, Derczynski, Pitenis, and \c{C}\"{o}ltekin}]{Zampieri2020}
Marcos Zampieri, Preslav Nakov, Sara Rosenthal, Pepa Atanasova, Georgi
  Karadzhov, Hamdy Mubarak, Leon Derczynski, Zeses Pitenis, and
  \c{C}a\u{g}r{\i} \c{C}\"{o}ltekin. 2020.
\newblock {SemEval-2020 Task 12: Multilingual Offensive Language Identification
  in Social Media (OffensEval 2020)}.
\newblock In \emph{Proceedings of SemEval}.

\end{thebibliography}

\end{document}